%% file: main.tex
\documentclass[conference]{IEEEtran}
\IEEEoverridecommandlockouts
\usepackage{cite}
\usepackage{verbatim}
\usepackage{amsmath,amssymb,amsfonts}
\usepackage{graphicx}
\usepackage{textcomp}
\usepackage{xcolor}
\usepackage{amssymb}
\usepackage{algpseudocode,algorithm}
\usepackage{algorithmicx}
\usepackage{booktabs}
\usepackage{outlines}
\usepackage{url}
\usepackage{adjustbox}
\usepackage{subcaption}

\makeatletter
\def\ps@IEEEtitlepagestyle{%
  \def\@oddfoot{\mycopyrightnotice}%
  \def\@evenfoot{}%
}
\def\mycopyrightnotice{%
  {\footnotesize - 978-1-6654-2476-9/21/\$31.00 ©2021 European Union\hfill}
  \gdef\mycopyrightnotice{}
}

\def\BibTeX{{\rm B\kern-.05em{\sc i\kern-.025em b}\kern-.08em
    T\kern-.1667em\lower.7ex\hbox{E}\kern-.125emX}}
\begin{document}

\title{Benchmarking Safety Monitors for Image Classifiers with Machine Learning

\thanks{The research leading to these results has received funding from the European Union’s Horizon 2020 research and innovation programme under the Marie Skłodowska-Curie grant agreement No 812.788 (MSCA-ETN SAS). This publication reflects only the authors’ view, exempting the European Union from any liability. Project website: http://etn-sas.eu/.}
}

\author{\IEEEauthorblockN{Raul Sena Ferreira}
\IEEEauthorblockA{
\textit{LAAS-CNRS, University of Toulouse,}
Toulouse, France \\
rsenaferre@laas.fr}
\\
\IEEEauthorblockN{Jeremie Guiochet}
\IEEEauthorblockA{\textit{LAAS-CNRS, University of Toulouse},
Toulouse, France \\
jeremie.guiochet@laas.fr}
\\
\and
\IEEEauthorblockN{Jean Arlat}
\IEEEauthorblockA{\textit{LAAS-CNRS, University of Toulouse},
Toulouse, France \\
jean.arlat@laas.fr}
\\
\IEEEauthorblockN{Helene Waeselynck}
\IEEEauthorblockA{\textit{LAAS-CNRS, University of Toulouse},
Toulouse, France \\
helene.waeselynck@laas.fr}

}

\maketitle

\input{sections/abstract}


\input{sections/1}
\input{sections/2}
\input{sections/3}
\input{sections/4}
\input{sections/5}

\bibliographystyle{IEEEtran}
\bibliography{references}

\end{document}

%% file: sections/abstract.tex
\begin{abstract}
High-accurate machine learning (ML) image classifiers cannot guarantee that they will not fail at operation. Thus, their deployment in safety-critical applications such as autonomous vehicles is still an open issue. The use of fault tolerance mechanisms such as safety monitors is a promising direction to keep the system in a safe state despite errors of the ML classifier. As the prediction from the ML is the core information directly impacting safety, many works are focusing on monitoring the ML model itself. 
Checking the efficiency of such monitors in the context of safety-critical applications is thus a significant challenge. 
Therefore, this paper aims at establishing a baseline framework for benchmarking monitors for ML image classifiers. Furthermore, we propose a framework covering the entire pipeline, from data generation to evaluation.
Our approach measures monitor performance with a broader set of metrics than usually proposed in the literature. Moreover, we benchmark three different monitor approaches in 79 benchmark datasets containing five categories of out-of-distribution data for image classifiers: class novelty, noise, anomalies, distributional shifts, and adversarial attacks.
Our results indicate that these monitors are no more accurate than a random monitor. 
We also release the code of all experiments for reproducibility.
\end{abstract}

%% file: sections/1.tex
\section{Introduction}\label{section-1}

Image classifiers based on machine learning (ML) are core components for many safety-critical autonomous applications, like autonomous vehicles. 
ML models make decisions based on trained past data.
A common way to validate them is by analyzing the discrepancy between the predicted values and the ground truth (labels) in a testing activity.
If the results are satisfactory, it goes to production.
However, even modern ML techniques, such as deep learning, can be wrong in their predictions even with 100\% confidence~\cite{gal2016dropout}.
This situation may lead to hazardous situations like the whole behavior of the system may rely on the ML decision. Therefore, it is now urgent to deploy techniques to increase the confidence in such ML classifiers. 
Among techniques coming from the dependability community \cite{avizienis2004basic}, fault tolerance is a technique that could be applied to deliver a correct service despite the occurrence of errors. Safety monitors (SM) keep the system in a safe state despite hazardous situations~\cite{machin2018smof}. 

Potentially, these situations may include errors from the ML model. 
Many recent works focus on monitors dedicated to the ML model surveillance. They broadly fall in three types of monitors: observation of the inputs of the ML model \cite{liu2020input}, its outputs \cite{HendrycksG16c} or from intermediate layers in case of deep neural networks (DNN)  \cite{cheng2019runtime, outsidethebox19}.
However, they are all based on the exploitation of the training data. Therefore, we refer to them as data-based monitors compared to safety monitors based on rules (or safety properties). Similar to uncertainties inherent to the use of ML, the confidence in such SM is an open issue. 

\begin{figure}
    \centering
    \includegraphics[width=4cm]{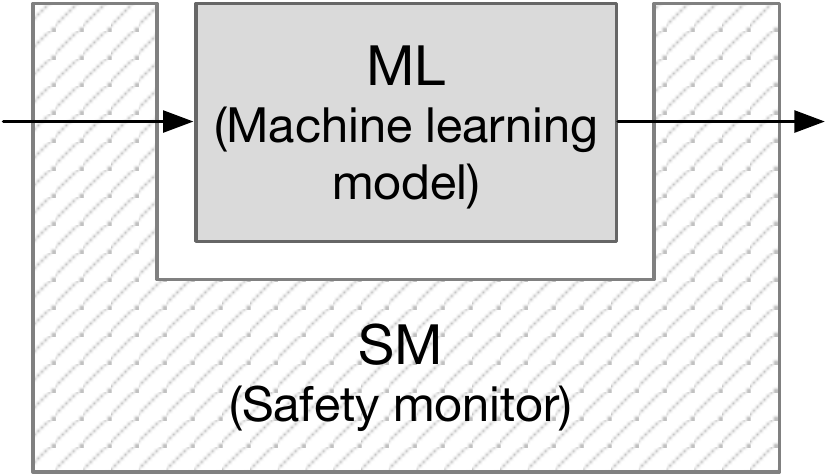}
    \caption{System under test composed of ML+SM}
    \label{fig:MLSM}
\end{figure}

Thus, it is essential to estimate how efficient such monitors are and if it is possible to include them in safety monitoring. This estimation should consider all potential situations leading to an error of the ML and should be based on metrics dedicated to measuring the monitor efficiency. Therefore, we focus on a framework for benchmarking such monitors, augmented with an additional primary mechanism to inhibit the decision in error detection. As presented in Figure~\ref{fig:MLSM}, the complete system under test is composed of an SM that surveil the ML. 

Our benchmark adapts and extends current metrics used in the ML community to estimate the SM detection performance at runtime, its impact on the system, and the overhead induced by the use of the monitor. 
Our main contributions are:

\begin{itemize}
    \item \textit{A new baseline framework for benchmarking SM for ML classifiers}. To the best of our knowledge, this is the first work that proposes an initial framework that benchmarks SM for ML components from different perspectives.
  
    \item \textit{A comprehensive benchmark experiment containing different \textbf{data-based} SM implementations, datasets, and results}. Our experiments reveal the advantages and drawbacks of the main modern data-based SM approaches for image classifiers built with ML. 
    We perform experiments on five challenging and important types of out-of-distribution data for image classifiers.
\end{itemize}

This work is organized as follows: in Section~\ref{section-2}, we provide background on ML classifiers, threats that may affect their results, and current safety monitoring approaches.
In Section~\ref{section-3}, we present an overview of our framework, along with its main objectives.
In Section~\ref{section-4} we present our experiments, datasets, and results of the application of our framework to three ML monitors.
Finally, in Section~\ref{section-5}, we present our conclusions.

%% file: sections/2.tex
\section{Background}\label{section-2}
This work focuses on monitoring ML image classifiers, especially when exposed to out-of-distribution data that can threaten their performance at runtime.
In the following subsections, we give a brief explanation of these concepts.

\subsection{Image classifiers with ML}
An ML classifier is a software component that uses an ML algorithm for identifying, given an observation, in which class it belongs \cite{alpaydin2020introduction}.
ML classifiers take an input, such as an image or a vector of numerical values, and outputs categorical values (classification) according to a category (class) previously seen during the training process. 
This component, called the ML model, is usually validated by analyzing the discrepancy between the classified values and the ground truth (labels).
If the results are satisfactory, it goes to production. 
These classifiers are integrated into many perception pipelines of autonomous systems. The prediction of the ML is at the very heart of the safety-critical functions of such systems (e.g., collision avoidance, path planning, and so on).

\subsection{Out-of-distribution data}
\label{sec:threats}
ML models tend to be biased to the training data \cite{nadeau2003inference}, resulting in a natural inability of a model to generalize 100\% of time even if all available data could be collected. Beyond this fundamental model generalization issue, there is another problem: data incompleteness.
That is, rare conditions tend to be underrepresented since the training data represents a subset of all real-world possibilities~\cite{shafaei2018uncertainty}.
It means that data represents the same target classes from training data, but that has different characteristics, different enough to threaten ML performance.
Such data is known as out-of-the-distribution data (OOD). 
In this paper, we consider five types of OOD data that can threaten an ML model at runtime: 

\begin{itemize}
    \item \textbf{Novel classes}: a scenario where new classes are introduced during runtime. For example, an ML model can be trained to identify dogs but fails to classify a new dog type not present in the training data~\cite{perera2019deep}.
    
    \item \textbf{Adversarial inputs}: a situation when ML fails with high confidence due to small input modifications~\cite{kurakin2016adversarial}.
    
    \item \textbf{Distributional shifts}: a condition that decreases the ML performance through time since the training dataset may differ from the real inputs. Such a situation is also known as a concept-drift ~\cite{ferreira2019amanda}. Examples of distributional shift include a change in class attributes such as dimension, physical characteristics, contrast, brightness, and other pixel-related variations in the image. 
    
    \item \textbf{Noise}: a possibility to receive deteriorated images due to small failures in exteroceptive sensors or unexpected environmental interference.
    
    \item \textbf{Anomalies}: a severe failure from exteroceptive sensors capable of severely corrupt images. However, contrary to noise, such failures corrupt data so that such images lose semantic value. It means that whatever the ML decision is, it can correspond to an inaccuracy. Examples of such images are black images, images with several shifted pixels, and so on.
\end{itemize}

Next, we detail the current approaches for monitoring ML.

\subsection{From SM to ML monitors}

Safety monitoring is a well-known dependability technique already used in embedded systems such as robots or autonomous vehicles.
This approach is generally based on a system model or the environment and on properties they should guarantee.
We will use rule-based monitors for such approaches (in opposition to data-based monitors presented in the following sub-section).  
Some techniques for monitoring safety components, consider this component as a \textbf{black-box}\cite{machin2018smof}. It is possible to apply this approach to systems, including ML algorithms. However, as they are considered black-box, the fact that it is an ML-based component has no real impact on the SM design.
For example, safety rules can be implemented in an SM  to verify if a vehicle can completely stop before reaching an obstacle \cite{ozguner2007systems} or to avoid a collision~\cite{al2017safety}.
In both cases, external (exteroceptive) sensors, such as distance sensors, are observed along with internal (proprioceptive) sensors, such as speed, to evaluate if a safety property is violated. It means that it was possible in these situations to have a redundant mechanism to monitor the ML.

However, when ML is in charge of image classification, in most cases, no redundant classification mechanism could be developed to monitor the ML. Indeed, it is usually impossible to use external sensors and external software not based on ML to confirm or invalidate an ML prediction. 

One possible direction is then to verify some assertions at the ML levels, for instance, with the use of model assertions\cite{kang2020model}.
This technique is an adaptation of the classical program assertions to monitor and improve ML models.
The idea is to verify inputs/outputs that indicate when errors may be occurring in the system.
For example, monitoring if an object flickers in and flickers out in the camera indicates a possible failure.
Even applying these techniques during design and operation cannot guarantee that a DNN decision is safe.
The reason is that, for some corner cases, ML outputs wrong decisions that cannot be verified by inspecting the code logic or the sensor values and lead to hazards\cite{dreossi2019compositional}.

We can deploy rule-based monitors when redundant observation sources are applied and safety properties are correctly expressed. However,  monitoring an ML component for vision is particularly complex to find a redundant source of observation. Moreover, it is not apparent to express a safety property at the level of an ML prediction. For these reasons, a current approach is to develop monitors based on the ML training data: the data-based monitors. 

\subsection{Data-based monitors} \label{sec:databasedmon}
This approach is usually based on data instead of using a system model or specifications provided by the developers. However, it means that we develop the monitor itself using the same training data as the ML.  
We propose to classify such monitors in three categories: monitors based  on ML \textit{inputs, intermediate values, or outputs}:

\subsubsection{\textbf{Monitoring DNN inputs}}
An example of this approach is to adversarially train an ML model to reconstruct a noisy image to an image similar to a previous known class \cite{sabokrou2018adversarially}.
It means that a new image arrives at runtime, and this ML model tries to reconstruct this image to a known one from the set of known classes previously seen in the training phase. 
Suppose there is a considerable difference between the reconstructed image and the actual image. In that case, this actual image is from an unknown class, indicating possible OOD data.
A similar idea is tested in \cite{denouden2018improving}, in which the authors apply the Mahalanobis distance to improve the autoencoder reconstruction process in order to capture OOD samples better.

Another possible solution based on DNN inputs is to create a radius distance threshold calibrated during the training~\cite{liu2020input}.
The idea is to perturb the DNN inputs, observe the correct answer, and determine how considerable the distance is regarding the DNN decisions.
The advantage of this approach is that it does not need to inspect the internals of the DNN.
The drawback is that it tends to be biased to the training data.

\subsubsection{\textbf{Monitoring the DNN intermediate values}}
An example of this approach is to monitor the neuron patterns observed from the layers of the DNN.
Such an approach usually compares the recorded patterns of the DNN activation functions during the training with those during runtime.
After the standard training process, a runtime monitor is created by feeding the training dataset to the network to store the neuron on-off activation patterns using binary decision diagrams (BDD)~\cite{cheng2019runtime} or using a 2D projection\cite{outsidethebox19}. 
For example, this projection is an abstraction box built by maximum and minimum activation function values during training.
It inspects if the output of an activation function falls inside of this abstraction box during runtime after a new input pass through it.
If not, it raises the alarm considering this input as a novel input.

Besides, uncertainty quantification methods can be applied to the intermediate outputs of a DNN\cite{berend2020cats}, recognizing activation patterns that do not represent any data from the training dataset.
An uncertainty-based supervisor accepts input as a typical sample if its uncertainty is lower than some threshold.
Thus, a high threshold leads to many false negatives, and a low threshold leads to many false positives \cite{weiss2021fail}.

\subsubsection{\textbf{Monitoring the DNN outputs}}
For instance, an SM can monitor the values in the last layer of a DNN\cite{HendrycksG16c}, verifying the decision's confidence level.
However, it is unreliable since DNNs can output a wrong decision even with a high confidence level.
For avoiding this problem, Liang et al. proposed enhancing the reliability of out-of-distribution image detection in neural networks (ODIN) \cite{liang2018enhancing} by applying techniques that decrease the DNN confidence values.

Uncertainty quantification methods can also be applied directly to DNN outputs.
Some methods that can be used for this purpose are point-prediction networks, MC-dropout, and DNN ensembles.
However, for evaluating such methods, it is necessary to apply metrics beyond the traditional combination of false/true positives/negatives, such as S-1 score \cite{weiss2021fail}.

\subsection{Benchmarking SM for ML-based components}
Despite several works for monitoring ML at runtime, just a few try to establish better benchmark methodologies.
One of these works proposes a less biased evaluation of out-of-distribution detectors, called OD-test \cite{shafaei2018less}.
Same as our work, the authors present a methodology that divides the data into three parts: training/validation, in-distribution testing set, and out-of-distribution testing set.
Henrikson et al. \cite{henriksson2019towards} also proposed a framework for benchmarking DNN monitors, using six OOD datasets and seven metrics.

Nevertheless, our work differs from the two related works presented above in three different aspects. 
Firstly, we propose a methodology that covers the entire pipeline, from data generation to evaluation.
Secondly, we apply metrics beyond the traditional accuracy and positive/negative rates, with additional statistical analysis in the results.
Finally, we perform tests using five different types of OOD data across a greater amount of generated datasets. 
Next, we present our framework.

%% file: sections/3.tex
\section{A baseline methodology for benchmarking SM for ML classifiers}\label{section-3}

\begin{figure}[htb!]
\centering
\includegraphics[width=0.95\linewidth]{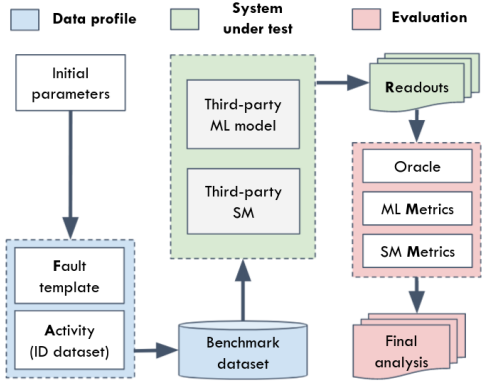}
\caption{A high-level overview of the framework.}
\label{fig:architecture}
\end{figure}

The proposed framework is an adaptation of the FARM \cite{arlat1990fault} methodology.
FARM is a methodology for fault injection for dependability validation.
It uses the input domain of a target system as a set of faults $F$ and a set of activations $A$ that specifies the domain to test the target system.
The output domain refers to a set of readouts $R$ used for posterior evaluation with a set of metrics $M$.
Our framework is divided into three modules, as illustrated in Figure \ref{fig:architecture}.

After setting some initial parameters such as type of data generation, the amount of data to be tested, which ML models and SM to include, starts with the first module, called the data profile.
This module generates benchmark datasets in the next module, called system under test (SUT).

The SUT is responsible for testing the ML and the SM performance, generating the results (readouts) at the end of the process. Readouts are inputs for the evaluation module.

The evaluation module is responsible for analyzing the readouts through several metrics for different components of the system.
It provides the final analysis for each type of readout.
Next, we explain each module and its components.

\subsection{Data profile}
This is the first module and it is composed of two items:

\begin{itemize}
    \item \textbf{Activity}: it contains in-distribution (ID) data, that is, instances from a distribution known by the ML.
    
    \item \textbf{Fault template}: it contains rules for generating out-of-distribution (OOD) data presented in Section II.B.
\end{itemize}

The benchmark dataset receives ID and OOD data to test how the ML and the SM behave when exposed to expected and unexpected data. 
Even though all types of OOD can be part of a unique benchmark dataset, we generated datasets divided by each category of OOD data.

An essential premise is that the generated benchmark dataset is not applied to build or validate the SM or train the ML model. 
This premise guarantees unbiased experiment results and is more realistic since, in real scenarios, there are no guarantees about which type of data will arrive at runtime.

\subsection{System under test}
The system under test (SUT) module receives the benchmark dataset as input and outputs readouts for each component.
To validate our framework, we chose ML models that classify images and one SM for each of the three categories mentioned in Section 2. 
Therefore, this module is composed of two main items:
a third-party \textbf{ML model} already trained; and a third-party \textbf{safety monitor} containing a detection mechanism.
We call the ML and the SM third-party components since they can be built aside from the framework.

This module simulates a stream of images randomly ordered.
We apply datasets with random images since the image datasets available in the literature use such a setting.

%
%
%

The simulation works as a stream of images coming from the benchmark dataset set $D$ that contains images $X$ and labels $y$, arriving at one of each at a time. 
The ML model receives this image and makes a $\hat{y}$ classification. 

Next, the system triggers the SM, which checks a set of predefined properties. 
These properties can be the ML’s classification $\hat{y}$ with the associated confidence level, intermediate layers, the input $X$, or even a combination of two or more of these properties. 
In the case of this pseudo-code example, the SM inspects the ML model’s internal properties (DNN’s hidden layer) during the classification, as recently suggested in the literature \cite{cheng2019runtime, outsidethebox19}.

After the inspection, the SM detector mechanism raises a detection alarm or not ($\hat{m}$). 
For this work, we consider a simple reaction strategy for the SM. 
If an alarm is raised, it invalidates the ML classification result; otherwise, it accepts the ML classification. 
Next, we observe whether such intervention produced a desirable outcome for the system or not. We call it \textit{overall detection} $\hat{s}$.
A desirable outcome means that canceling an ML output/agreeing with it was beneficial to the SUT. This result is independent of whether the SM-specific detection was correct or not in the OOD detection task. 
The criteria for considering these two dimensions of the detection (specific and overall) correct or not is discussed in the following subsection. 
This process continues until the end of the stream. 
Thus, the readouts are: the ground truth $y$, the ML classification $\hat{y}$, the SM specific detection $\hat{m}$, and its overall detection $\hat{s}$.
They are divided into three different categories:

\begin{enumerate}
    \item \textbf{ML readouts}: it contains classification (e.g., class number), confidence value from the last layer, and intermediate values from hidden layers.
    \item \textbf{SM readouts}: it contains two types of detection: a) specific (e.g., it outputs 1 for OOD detection; 0 otherwise); b) overall (e.g., after OOD detection, it cancels the ML decision (1); 0 otherwise).
    \item \textbf{General readouts}: it contains general outputs from all components above, such as processing time and memory.
\end{enumerate}

As can be noted, the SM readouts contain results in two dimensions: specific and overall.
Considering two dimensions for detection is more realistic and complete than just analyzing the detection rate for the OOD data. 
The reason is that while the SM tries to detect OOD data, it also can avoid that the ML gives an answer that is different from the ground truth.
The opposite is also true. As a result, the SM can incorrectly detect ID data as OOD data, hindering the correct decisions of ML. 
All possible situations are given in the next Subsection.

\subsection{Evaluation}
This module receives the readouts as inputs containing false positives/negatives and true positives/negatives regarding OOD data detection's specific task and the overall task of avoiding an unsafe outcome. 
This module evaluates two main aspects:

\begin{enumerate}

    \item \textbf{The SMs performance:} the objective is to assess the SM results regarding the specific task of detecting OOD data.
    This evaluation’s inputs are SM OOD detection (e.g., raises/does not raise the alarm).
    
    \item \textbf{The overall impact of the SM:} the objective is to determine if the SM improves or worsens the overall SUT accuracy. 
    We analyze the SUT using the ML alone (baseline) and ML with the SM.
    This evaluation’s inputs are the overall decisions made by the SUT with ML alone and ML with SM, processing time, and memory.
    
    \item \textbf{Time and memory overhead induced by SMs:} the objective is to investigate the memory and processing time efficiency of these methods during runtime.

\end{enumerate}

This module contains two major items: the \textbf{oracle}, and the \textbf{metrics for ML and SM}.
The oracle determines positive and negative data and whether a test has passed or failed.
Thus, the oracle considers a correct specific detection when the SM correctly detects OOD data.
Besides, the oracle considers a correct overall detection for the SUT when the SM correctly avoids a wrong ML classification for ID or OOD data. 
The oracle takes into consideration the following scenarios to determine whether a readout is correct or incorrect:

\begin{outline}

    \1 \textbf{ID data arrives in the stream}
    
        \2 \textbf{If the SM detects OOD data:}
        it means a \emph{false positive for the specific task} of OOD detection.
    
            \3 \textbf{If the ML classification is equal to the ground truth:} it means a \emph{false positive for the overall task} of avoiding a failure since the SM intervened without necessity.
            
            \3 \textbf{If the ML classification is different than the ground truth:} it means a \emph{true positive for the overall task} of avoiding a failure since it canceled the ML misclassification for ID data.
            
        \2 \textbf{If the SM does not detect OOD data:}
        it means a \emph{true negative for the specific task} of OOD detection.
        
            \3 \textbf{If the ML classification is equal to the ground truth:} it means a \emph{true negative for the overall task} of avoiding a failure since the ML gave a correct classification even though the SM did not detect OOD data.
            
            \3 \textbf{If the ML classification is different than the ground truth:} it means a \emph{false negative for the overall task} of avoiding a failure since the ML misclassified the OOD data and the SM did not detect it, which could cancel a wrong ML output.
        
    \1 \textbf{OOD data arrives in the stream}
    
        \2 \textbf{If the SM detects OOD data:}
        it means a \emph{true positive for the specific task} of OOD detection.
        
            \3 \textbf{If the ML classification is equal to the ground truth:} it means a \emph{false positive for the overall task} of avoiding a failure since the ML gave a correct classification but the SM wrongly intervened.
            
            \3 \textbf{If the ML classification is different than the ground truth:} it means a \emph{true positive for the overall task} of avoiding a failure since the ML gave an incorrect classification and the SM correct canceled the ML decision.
            
        \2 \textbf{If the SM does not detect OOD data:}
        it means a \emph{false negative for the specific task} of OOD detection.
        
            \3 \textbf{If the ML classification is equal to the ground truth:} it means a \emph{true negative for the overall task} of avoiding a failure since the SM detection for OOD data also avoided an ML misclassification.
            
            \3 \textbf{If the ML classification is different from the ground truth:} it means a \emph{false negative for the overall task} of avoiding a failure since the SM detection for OOD data also avoided an ML misclassification.
            
\end{outline}

The only exception for the above rules is for novelty class detection.
When OOD data arrives in the stream, if the SM correctly detects OOD data, it will always be interpreted as true positive since the SM always correctly cancel the ML classification, independently of the ground truth. 
Conversely, if the SM does not detect OOD data, it will always be a false negative. If the SM lets the ML deal with a class that it was not trained in before, it can be regarded as unsafe. In order to evaluate these situations, we have selected in the literature a set of metrics that are pertinent for our study. Hence, we apply seven metrics: 

\begin{itemize}
    \item \textit{Matthews correlation coefficient (MCC)}: it ranges from -1 if (ML or SM always wrong) passing through 0 (ML or SM is accurate as random) to 1 (ML or SM always right). This metric is more reliable than traditional metrics such as accuracy. It yields a high score only if the ML or SM can be assertive in all of the four confusion matrix categories \cite{chicco2020advantages}. In contrast, accuracy only considers the portion of the right answers. 
    In a scenario in which the number of a class is 80\%, the accuracy could yield a score of 80\% for an SM that did not detect other classes.
    
    \item \textit{False positive rate (FPR)}: also known as type-I error. It indicates how many false alerts the SM raises for the task to detect OOD data.
    High values means the SM indicates a problem wherein there are not in most cases.
    
    \item \textit{False negative rate (FNR)}: also known as type-II error. It indicates how often the SM misses detecting OOD data.
    If this value is high, it indicates that the SM does not recognize the difference between the ID and OOD data.
    
    \item \textit{Precision and recall (Pr and Re)}: the fraction of correct detection and the fraction of available OOD data. These metrics help to indicate how well the SM detected OOD data through the benchmark dataset.
    
    \item \textit{Micro-F1}: harmonic mean (global) for the prediction x recall.
    It helps to assess the quality of multi-label binary problems, which makes them suitable to be applied in the SUT evaluation.
    
  %
    \item \textit{Critical difference diagram}: it is applied for all MCC results through all datasets. It shows how statistically different are the results between the SMs.
    
    
\end{itemize}

Next, we detail each component's choice and its parameters for the three modules, and the benchmark results. 

%% file: sections/4.tex
\section{Experiments and results}\label{section-4}

\subsection{Data profile}
We use three datasets as the \emph{activity} of our framework:
\begin{itemize}
    \item GTSRB \cite{sermanet2011traffic}: German traffic signs with 43 classes, with 39200 instances for training and 12600 for testing.
    \item BTSC \cite{jain2019novel}: Belgium traffic signs with 62 classes, divided into 7000 images for training and testing.
    \item CIFAR-10 \cite{krizhevsky2010convolutional}: ten general classes (e.g., dog, car ...), with 50000 instances for training and 10000 for testing.
\end{itemize}

Next, we apply a \emph{fault template} for the five classes of OOD presented in section~\ref{sec:threats}: novelty, anomaly, distributional shift, noise, and adversarial inputs. 
For adversarial inputs generation, we apply the fast gradient signed method (FGSM) \cite{dong2017discovering}.
For the noise and distributional shift, we generate 19 different transformations with two types of intensity varying from 1 to 5 (e.g., snow (1) = image with a bit of snow; snow (5) = heavy snow) \cite{hendrycks2019benchmarking}.
All transformations were applied over CIFAR-10 and GTSRB datasets.
The benchmark datasets are composed of a transformed version of the ID datasets that we previously applied for training the ML algorithm and building the SM.
Thus, we use 20\% of the ID data for the benchmark dataset without transformation. The entire original dataset is transformed into a specific variation.
For obvious reasons, this 20\% of ID data was also not applied to the ML training.

For novelty class detection, the fault template applies one dataset as an ID dataset and another dataset with new classes as OOD data, resulting in three benchmark datasets:
\begin{itemize}
    \item GTSRB as ID data, and  BTSC as OOD data: this combination tests whether the SM can distinguish new classes that have similar characteristics to the known ones.
    \item GTSRB as ID data, and  CIFAR-10 as OOD data: this combination tests whether the SM can distinguish new classes that are very different from the known ones.
    \item CIFAR-10 as ID data, and  GTSRB as OOD data: this combination tests the same as the aforementioned combination. However, since the ID data is different, the ML and the SM are built with different data. Hence, this permutation produces different outcomes. 
\end{itemize}

In total, we produced 79 benchmark datasets.
More details can be found in the results section of our repository \cite{BEN}.

\subsection{System under test}
We simulate a scenario in which the SM has to detect OOD data by checking one RGB-colored image at a time in a randomly ordered stream of images.
Once the SM makes a detection, it cancels the ML classification since this classification is potentially wrong. If nothing is detected, the ML classification is accepted.

For the ML model, we use a LeNet \cite{lecun2015lenet} since it is a simple and traditional convolutional neural network (CNN) algorithm. It contains around 100,000 parameters and 128 neurons in the last hidden layer. 
For the monitors (noted as SM), we use one method of each of the three different strategies of data-based SM (presented in section~\ref{sec:databasedmon}) for monitoring a DNN model:
\begin{itemize}
    \item \textit{DNN inputs}: adversarially learned one-class classifier for novelty detection (ALOOC) \cite{sabokrou2018adversarially}.
    This method learns how to reconstruct each class during the training phase. It receives an image during operation, tries to reconstruct this image to the known class, and analyzes the loss error resulting from this reconstruction.
    If this loss error surpasses a safe threshold, ALOOC flags it as OOD.
    \item \textit{DNN intermediate values}: outside-of-the-box abstraction (OOB) \cite{outsidethebox19}.
    This method projects a 2D-box region from the activation function values (RELU) from the hidden layers of the DNN during the training phase.
    At runtime, it receives an image and projects a point from it built from the outputs of the activation function from the same DNN layer used during the training. 
    If this point falls outside the 2D box, OOB will flag this image as OOD.
    \item \textit{DNN outputs}: detector of out-of-distribution images in neural networks (ODIN) \cite{liang2018enhancing}.
    This method learns how to balance the confidence values outputted along with the DNN decision.
    It uses these values from the last layer of DNN during training and applies a method known as temperature scaling, which will scale the confidence values to a new threshold for considering an image as OOD.
    During the operation, ODIN verifies if the DNN confidence value over an incoming image respects the threshold. If not, ODIN will flag this image as OOD.
\end{itemize}

\subsection{Parameter's evaluation and choice}
Next, we present the chosen parameters for the SMs.
To avoid biased results, all the initial parameters were tuned using only the ID dataset and are summarized in Table \ref{table_parameters}.

\begin{table}[htb!]
\centering
\caption{SM parameters tested in the experiments.}
    \begin{tabular}{lll}
    \toprule
    
    \textbf{SM} & \textbf{Parameter name}& \textbf{Parameter values} \\
    
    \toprule
    
    ALOOC & Optimizer & ADAM, RMSProp \\
         & Epoch  & 200 \\
         & Loss threshold & average per class*\\
    
    \toprule
    
    OOB & $\gamma$ & 0, 0.1, 0.35 \\
    & \# of clusters & 0, 3, 5*, 17* \\
    & Dimensionality reduction & simple, PCA*, ISOMAP*\\
    
    \toprule
    
    ODIN & Temperature & 1000 \\
        & Magnitude & 0.0014, 0.0025 \\
        & Confidence threshold & 0.0237*, 0.1007* \\
        
    \bottomrule
    \end{tabular}
\label{table_parameters}
\end{table}

We explored the original parameters for each SM, and we also found/tested new values/methods not mentioned in their original papers (marked with *).
For ALOOC, the authors did not investigate the influence of the optimizers in the original paper. Thus, we tested two different optimizers: ADAM \cite{zhang2018improved} and RMSProp \cite{wichrowska2017learned}.
We also analyzed the best model for each class through different epochs.
In general, all the models achieved a better convergence around 200 epochs. 
Since the threshold value for considering whether a class is considered OOD does not indicate in the original paper, we considered the average reconstruction loss for each class during the training.


For the OOB method, we test the parameter responsible for enlarging the size of boxes ($\tau$) with the same range of values proposed by the original paper. 
We applied the same number of clusters as suggested by the original paper (no clusters or three clusters). However, we also tested with a possible optimal number of clusters.
The possible optimal number of clusters K for the K-means algorithm contained in the OOB method is chosen through the Elbow analysis \cite{kodinariya2013review}.
Thus, the best K value was 5 and 17 for CIFAR-10 and GTSRB, respectively. 
Finally, we also tested three different approaches for the 2D-dimensionality reduction parameter: simple projection (proposed in the original paper), PCA \cite{yang2004two}, and ISOMAP \cite{balasubramanian2002isomap}.
ISOMAP is a popular nonlinear dimensionality reduction method, and PCA is also a popular method, but linear.
With these two suggested methods, we can analyze if the choice of the dimensionality reduction methods influences the outcomes.

Regarding ODIN, we set the temperature (1000) and the magnitude parameters (0.0025 and 0.0014, for GTSRB and CIFAR-10, respectively) as suggested by the authors.
However, we chose the confidence thresholds for determining OOD data by selecting the lower confidence value outputted from the method when exposed to the training data.
In this case, 0.0237 and 0.1007, for GTSRB and CIFAR-10, respectively.
We had to assume a confidence value threshold since it would not be possible to use ODIN as an SM without it.

\subsection{Results}
\subsubsection{\textbf{SMs performance}}
we use the positive/negative data and the evaluation metrics as mentioned in Section \ref{section-3}.C.
Since we measure the detection, there is no necessity to include the measurements for the ML.
Best results are written in \textbf{bold}.

Table \ref{table_novelty} shows these results for \emph{novelty class}: GTSRB as ID dataset and BTSC as OOD dataset; CIFAR-10 as ID and GTSRB as OOD; and GTSRB as ID and CIFAR-10 as OOD.
The first benchmark dataset has the challenge of having ID and OOD data with a similar domain, while the other two have very different classes from each other.

\input{sections/tables/novelty_class_tables}

The MCC results indicate that the SM is sometimes slightly better than a random classifier (MCC $\approx$ 0).
We observe that these methods' weakness is in the high amount of false positives, yielding a borderline MCC performance.
According to the results, just ALOOC obtained a false-positive rate of around 50\%. However, it had many false negatives, which can be considered worse depending on the scenario.
In general, all SM had a poor performance due to the unreliable nature of DNN confidence values, and the high nonlinear nature of activation functions in ODIN and OOB, respectively.




Next, in Table \ref{table_adversarial}, we show the same analysis but using CIFAR-10 and GTSRB after being modified through an \emph{adversarial attack} known as fast gradient sign method (FGSM).

\input{sections/tables/adversarial_tables}

According to the results, for CIFAR-10, the methods achieved negative values for MCC, which indicates performance worse than a random classifier. 
For GTSRB, the results are slightly better, but the MCC values can be considered as flawed as a random classifier.
Despite the good values of micro-F1 for ODIN and ALOOC, in the CIFAR-10 and GTSRB datasets, respectively, the rate of false negatives was high.
It means that the ML classifier did not give the correct prediction, and the SM did not detect the attack.

Next, in Table \ref{table_dist_shift}, we show the results for CIFAR-10 and GTSRB datasets with different types of \emph{distributional shift}.
Here, we see a surprising but negative outcome: OOB suffers from a lot of false positives. In contrast, ODIN suffers from many false negatives.
It means that they always tend to miss a detection or to say that everything is OOD data.
ALOOC got a fair MCC value for CIFAR-10 with intense fog and a good MCC value for GTSRB with heavy snow.

\input{sections/tables/dist_shift_tables}

Another interesting result for ALOOC is that despite the low MCC performance on the other benchmark datasets, it got relatively good micro-F1 results.
It highlights that this method tends to have a good amount of false negatives but rarely gives a wrong output when detection is signaled.
Again, methods based on inspecting inputs seem to be more effective in detecting changes in the pixel values distribution.

\begin{figure}[!htb]
\centering
\includegraphics[width=0.8\linewidth]{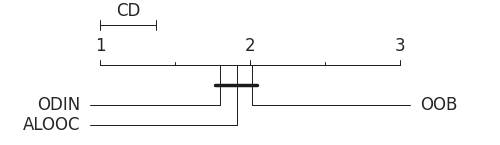}
\caption{Critical difference diagram between the three methods.}
\label{fig:stats_cd}
\end{figure}

Once we evaluate the methods in all datasets, we investigate how their results statistically differ from each other.
The reason is to observe if there is an SM that is better than the others.
We applied a Friedman test with Nemenyi posthoc with 95\% the confidence level for all methods ranks through every dataset regarding the MCC metric. 
From that, we build a critical difference diagram illustrated in Figure \ref{fig:stats_cd}.

The result shows that ODIN achieved the best results in more benchmark datasets than the other two SM. However, there is no statistical difference between the SMs.
It means that regarding the MCC results, there is no best SM method.
Next, we evaluate how much the SM impacts the SUT when it works along with the ML classifier.

\subsubsection{\textbf{Overall impact of the SM}}
As previously explained, the objective is to evaluate how much the SM impacts the SUT when it works along with the ML classifier.
Table \ref{table_1} shows the overall SM's impact in the SUT when using GTSRB or CIFAR-10 regarding ID data.
The results are expressed as MCC values. To evaluate the SM, we also use a percentage of relative change showing how much better/worse it performed compared to the baseline (ML alone).

\begin{table}[htb!]
\centering
\caption{MCC values for SUT with/without SM for GTSRB or CIFAR-10 as ID dataset.}
\label{table_1}
\begin{tabular}{lll}
\toprule
Method &                GTSRB &           CIFAR-10 \\
\midrule
    ML alone &     \textbf{0.96} &  \textbf{0.74}\\
    ML + ALOOC &  0.51 &   0.53 \\     
    ML + OOB &  0.64  &   0.61  \\
    ML + ODIN &   0.50 &  0.64  \\
\bottomrule
\end{tabular}
\end{table}

In the GTSRB dataset, the best values for the ALOOC method were obtained when using the ADAM optimizer. In contrast, the OOB method obtained the best results when using 3 clusters, ISOMAP, and an enlargement factor of 0.1.
As mentioned earlier, for ODIN, the threshold was set to 0.0237.

For the CIFAR-10 dataset, the best optimizer for ALOOC was ADAM.
For OOB, the best parameters were: 5 clusters, ISOMAP, and an enlargement factor of 0.35.
For ODIN, we set the threshold to 0.1007.

As can be noted, all SM methods perform worse than the ML alone when exposed to ID data.
This result is expected due to the generalization power of ML models.
However, a reliable SM is also the one that can perform well when exposed to ID data, avoiding an ML misclassification or simply not raising a false alarm hindering a correct ML classification.

These results show the best SM's performance was obtained by the outside-of-the-box and ODIN, on GTSRB and CIFAR-10, respectively. 
Results indicate that the SM performed consistently better than a hypothetical random classifier ($MCC=0$).
However, they also show a significant gap between the performance of the ML alone and the SM for ID data, especially in the GTSRB that has four times more classes than CIFAR-10.
This result is interesting since the tested monitors use information from the ML model and build their hypothesis over ID data.

\subsubsection{\textbf{Time and memory overhead induced by SMs}}
We show in Table \ref{table_time} an example of the performance for ALOOC, OOB, and ODIN, respectively, in the three benchmark datasets for novelty class detection.
The values are expressed in seconds and represent the average time spent on the tasks of classification and monitoring on a single image. We used a computer with a Processor Intel(R) Core(TM) i7-9850H CPU @ 2.60GHz, and 32GB of memory.

Worth mentioning that we can have slight differences between the prediction time in the three methods since they use different deep learning libraries. ALOOC uses Keras, OOB uses Tensorflow, and ODIN uses PyTorch.
Besides, the SUT time also contains the remaining time from other small processes involved in the experiment, so the summing between the prediction and monitoring time is lower than the SUT time.

\begin{table}[htb!]
\centering
\caption{Time impact of SM per instance in seconds for novelty class detection.}
\label{table_time}
\begin{tabular}{llllll}
\toprule

Method & ML &           SM     &      SUT \\
\midrule
       ALOOC & 0.0070 (3.0\%) & 0.2217 (96.8\%) &   0.2288 \\
	OOB & 0.0021 (3.9\%)&  0.0529 (96.0\%)&  0.0551 \\
	ODIN & 0.0007 (3.0\%) &  0.0246 (96.9\%)&  0.0254 \\

\bottomrule
\end{tabular}
\end{table}

According to the results, ODIN is the fastest method. Except when using OOB with ISOMAP, all three methods needed no more than 0.07 seconds to spot OOD data. Such process time can be considered fast enough to be applied at runtime.
Besides, they can be optimized, potentially reducing this time.

However, the monitoring task is always slower than the prediction and can be responsible for 96\% of the necessary time for a SUT to do the task.
Since memory is also an important constraint, especially on embedded systems, we show in Table \ref{table_memory} the memory efficiency of the methods when performing a novelty class detection over the GTSRB dataset.

\begin{table}[htb!]
\centering
\caption{Memory size of ML, and SM in MB.}
\label{table_memory}
\begin{tabular}{lll}
\toprule

Method & ML &           SM   \\
\midrule
       ALOOC & 3.8 & 98.9  \\
	OOB & 4.3 & 6.4  \\
	ODIN & 2 & 1.5  \\

\bottomrule
\end{tabular}
\end{table}

As can be seen, ALOOC needs a considerable amount of memory or disk space due to the necessity of developing one monitor for each class.
For instance, even though ALOOC needs just 2.3MB per class, it is necessary almost 100MB to monitor all 43 classes contained in the GTSRB dataset.
For OOB, it is also necessary to build one monitor for each class.
However, the amount of memory needed is not huge since it uses just some stored arrays to make the boxes.
For ODIN, it is not necessary to build one SM for each class.
Moreover, since the algorithm needs to be applied during the training phase to collect the thresholds for the confidence values, it can use just a tiny amount of memory to do the inspection.

\subsection{\textbf{Threats to validity}}
In order to analyze and mitigate threats to the validity of the results, we present below a summary of arguments for external and internal validation.
\begin{itemize}
    \item \emph{External validity}: 
    All tested SMs were already compared to other methods in their original papers. They obtained the best results during the comparison.
    Besides, we also chose three different approaches of monitoring (inputs, intermediate values, outputs) that may cover an acceptable range of SM approaches based on data.
    However, it is worth mentioning that this study is a first benchmark work. Thus, other approaches may be published with better performances in the future.
    
    \item \emph{Internal validity}: 
    What could have to lead us to the wrong conclusions in our study?
    \begin{itemize}
        \item \emph{Bad parameters choices for the SM}: 
        aiming at improving the SM performance, we explored additional parameter values than those explored in their original papers.
        For instance, we applied different dimensionality reduction methods for the outside-the-box. 
        However, even though different dimensionality reduction methods can bring better results, they also introduce a high cost to time and memory performance.
        Regarding ALOOC and ODIN, we had to choose criteria to the threshold value for considering whether a class is considered out-of-distribution or not.
        This way, if one applies different criteria for these methods, they will result in different outcomes.
        However, such criteria had to be deducted since they are not contained in their original papers. 
        Besides, ODIN used OOD data to calibrate their monitor, what can be considered unrealistic. 
        Hence, in this work, all the parameters were chosen using just ID data.
        It makes the scenario more realistic and harder, which drastically decreases the SMs performance.
        
        \item \emph{Choice of datasets}: 
        even though we chose datasets widely applied in the computer vision literature, the choice of the amount of ID data accessible for building the ML classifier and the SM can influence both.
        Furthermore, the amount of ID and OOD data in the benchmark datasets also can influence the results.
        However, we followed traditional ways to divide data (e.g., 80/20 for training and test).
        Besides, the authors of OOB \cite{outsidethebox19}, and ALOOC \cite{sabokrou2018adversarially} test their methods considering the same dataset for ID and OOD data, which they test novelty detection by training the ML with 9 classes and omitting one, or training with 8 classes and omitting two, and so on. However, we in this work we test novelty detection using an entirely new dataset as OOD data along with a part of ID data (ex: CIFAR-10 + GTSRB).
        
        \emph{Domain of validity}:
        The setting CIFAR10-GTSRB constitutes two different datasets that do not have the same validity domain \cite{riccio2020model} (i.e., GTSRB is not only out-of-distribution but also out of the validity domain).
        Thus, the task of detecting OOD data should be less complicated in this scenario.
        However, the SM methods continued to be inefficient.
        
    \end{itemize}
    
\end{itemize}

%% file: sections/tables/novelty_class_tables.tex
\begin{table}[htb!]
\centering
\caption{Comparing data-based monitors for \textbf{novelty class}.}
\resizebox{\columnwidth}{!}{%
\label{table_novelty}
\begin{tabular}{llllllll}
\toprule
    \textbf{Variation} & \textbf{SM} &            \textbf{MCC} &            \textbf{FPR} &           \textbf{FNR} &      \textbf{Pr} &        \textbf{Re} &       \textbf{F1} \\
\midrule
    GTSRB-BTSC    &
    \begin{tabular}{@{}c@{}} ALOOC \\ OOB \\ ODIN \end{tabular} &  
    \begin{tabular}{@{}c@{}}0.01 \\ \textbf{0.23} \\ 0.03\end{tabular} &    
    \begin{tabular}{@{}c@{}}\textbf{0.50} \\ 0.61 \\ 0.99\end{tabular}&          
    \begin{tabular}{@{}c@{}}0.49 \\ 0.11 \\\textbf{0.0} \end{tabular}&            
    \begin{tabular}{@{}c@{}}0.17 \\ \textbf{0.24} \\ 0.16 \end{tabular}&          
    \begin{tabular}{@{}c@{}}0.51 \\ 0.90 \\ \textbf{1.0}\end{tabular}&  
    \begin{tabular}{@{}c@{}}\textbf{0.57} \\ 0.52 \\0.06\end{tabular}\\
    
\midrule

   CIFAR10-GTSRB &
   \begin{tabular}{@{}c@{}} ALOOC \\ OOB \\ ODIN \end{tabular} &  
   \begin{tabular}{@{}c@{}} 0.02 \\ 0.11 \\ \textbf{0.23} \end{tabular} &           
   \begin{tabular}{@{}c@{}}0.63 \\  0.72 \\ \textbf{0.61} \end{tabular}&            
   \begin{tabular}{@{}c@{}}0.34 \\ 0.16 \\ \textbf{0.10} \end{tabular}&   
   \begin{tabular}{@{}c@{}}0.18 \\ 0.20 \\ \textbf{0.24} \end{tabular}&           
   \begin{tabular}{@{}c@{}}0.66 \\ 0.84 \\ \textbf{0.9} \end{tabular}&   
   \begin{tabular}{@{}c@{}}0.47 \\ 0.41 \\ \textbf{0.52} \end{tabular}     \\     

\midrule

    GTSRB-CIFAR10 & 
   \begin{tabular}{@{}c@{}} ALOOC \\ OOB \\ ODIN \end{tabular} &  
   \begin{tabular}{@{}c@{}}0.05 \\ \textbf{0.15} \\ 0.07 \end{tabular}&  
   \begin{tabular}{@{}c@{}}\textbf{0.56} \\ 0.79 \\ 1.0 \end{tabular}&   
   \begin{tabular}{@{}c@{}}0.37 \\ 0.09 \\ \textbf{0.02} \end{tabular}&           
   \begin{tabular}{@{}c@{}}0.81 \\ \textbf{0.82} \\ 0.17 \end{tabular}&  
   \begin{tabular}{@{}c@{}}0.63 \\ 0.91 \\ \textbf{0.98} \end{tabular}&           
   \begin{tabular}{@{}c@{}}0.63 \\ \textbf{0.74} \\ 0.06 \end{tabular}    \\       
\bottomrule
\end{tabular}
}
\end{table}

%% file: sections/tables/adversarial_tables.tex
\begin{table}[htb!]
\centering
\caption{Comparing data-based monitors for CIFAR-10 and GTSRB datasets with a \textbf{adversarial attack}.}
\resizebox{\columnwidth}{!}{%
\label{table_adversarial}
\begin{tabular}{llllllll}
\toprule
\multicolumn{8}{c} {CIFAR-10}\\
\toprule
Variation &                                             Method &                                                MCC &                                                FPR &                                                FNR &                                          Pr &                                             Re &                                           F1 \\
\midrule
     FGSM &  \begin{tabular}{@{}c@{}}ALOOC \\ OOB \\ ODIN\end{tabular} &  
     \begin{tabular}{@{}c@{}}-0.23 \\ -0.13 \\ \textbf{0.06}\end{tabular} &  
     \begin{tabular}{@{}c@{}}0.89 \\ 0.92 \\ \textbf{0.14}\end{tabular} &  
     \begin{tabular}{@{}c@{}}0.29 \\ \textbf{0.16} \\ 0.81\end{tabular} &  
     \begin{tabular}{@{}c@{}}0.28 \\ 0.31 \\ \textbf{0.37}\end{tabular} &  
     \begin{tabular}{@{}c@{}}0.71 \\ \textbf{0.84} \\ 0.19\end{tabular} &  
     \begin{tabular}{@{}c@{}}0.25 \\ 0.24 \\ \textbf{0.62}\end{tabular} \\
\bottomrule
\bottomrule
\multicolumn{8}{c} {GTSRB}\\
\toprule
Variation &                                             Method &                                                MCC &                                                FPR &                                                FNR &                                          Pr &                                             Re &                                           F1 \\
\midrule
     FGSM &  \begin{tabular}{@{}c@{}}ALOOC \\ OOB \\ ODIN\end{tabular} &  
     \begin{tabular}{@{}c@{}}\textbf{0.19} \\ -0.01 \\ 0.11\end{tabular} &  
     \begin{tabular}{@{}c@{}}\textbf{0.22} \\ 1.0 \\ 0.92\end{tabular} &  
     \begin{tabular}{@{}c@{}}0.59 \\ \textbf{0.0} \\ 0.02\end{tabular} &  
     \begin{tabular}{@{}c@{}}0.44 \\ \textbf{0.31} \\ 0.34\end{tabular} &  
     \begin{tabular}{@{}c@{}}0.41 \\ \textbf{1.0} \\ 0.98\end{tabular} &  
     \begin{tabular}{@{}c@{}}\textbf{0.66} \\ 0.14 \\ 0.26\end{tabular} \\
\bottomrule
\end{tabular}
}
\end{table}


%% file: sections/tables/dist_shift_tables.tex
\begin{table}[htb!]
\centering
\caption{Comparing data-based monitors for CIFAR-10 and GTSRB with different types of \textbf{distributional shift}.}
\resizebox{\columnwidth}{!}{%
\label{table_dist_shift}
\begin{tabular}{llllllll}
\toprule
\multicolumn{8}{c} {CIFAR-10}\\
\toprule
Variation &                                             Method &                                                MCC &                                                FPR &                                                FNR &                                          Pr &                                             Re &                                           F1 \\
\midrule
     rotated &  \begin{tabular}{@{}c@{}}ALOOC \\ OOB \\ ODIN\end{tabular} &  
\begin{tabular}{@{}c@{}}0.0 \\ \textbf{0.02} \\ -0.1\end{tabular} &  
\begin{tabular}{@{}c@{}}\textbf{0.0} \\ 1.0 \\ 0.32\end{tabular} &  \begin{tabular}{@{}c@{}}0.29 \\ \textbf{0.0} \\ 0.81\end{tabular} &  \begin{tabular}{@{}c@{}}\textbf{1.0} \\ 0.14 \\ 0.09\end{tabular} &  \begin{tabular}{@{}c@{}}0.71 \\ \textbf{1.0} \\ 0.19\end{tabular} &  \begin{tabular}{@{}c@{}}\textbf{0.83} \\ 0.04 \\ 0.66\end{tabular} \\

\midrule

 snow  (5) &  \begin{tabular}{@{}c@{}}ALOOC \\ OOB \\ ODIN\end{tabular} &  
 \begin{tabular}{@{}c@{}}-0.01 \\ 0.0 \\ \textbf{0.14}\end{tabular} &  
 \begin{tabular}{@{}c@{}}0.42 \\ 1.0 \\ \textbf{0.08}\end{tabular} &  \begin{tabular}{@{}c@{}}0.59 \\ \textbf{0.0} \\ 0.81\end{tabular} &  \begin{tabular}{@{}c@{}}0.14 \\ 0.14 \\ \textbf{0.29}\end{tabular} &  \begin{tabular}{@{}c@{}}0.41 \\ \textbf{1.0} \\ 0.19\end{tabular} &  \begin{tabular}{@{}c@{}}0.62 \\ 0.04 \\ \textbf{0.8}\end{tabular} \\
 
 \midrule
 
 fog  (5) &  \begin{tabular}{@{}c@{}}ALOOC \\ OOB \\ ODIN\end{tabular} &  
 \begin{tabular}{@{}c@{}}\textbf{0.47} \\ 0.0 \\ -0.01\end{tabular} &  
 \begin{tabular}{@{}c@{}}\textbf{0.03} \\ 1.0 \\ 0.21\end{tabular} &  \begin{tabular}{@{}c@{}}0.59 \\ \textbf{0.0} \\ 0.81\end{tabular} &  \begin{tabular}{@{}c@{}}\textbf{0.68} \\ 0.14 \\ 0.13\end{tabular} &  \begin{tabular}{@{}c@{}}0.41 \\ \textbf{1.0} \\ 0.19\end{tabular} &  \begin{tabular}{@{}c@{}}\textbf{0.88} \\ 0.04 \\ 0.73\end{tabular} \\
\bottomrule
\bottomrule
\multicolumn{8}{c} {GTSRB}\\
\toprule
Variation &                                             Method &                                                MCC &                                                FPR &                                                FNR &                                          Pr &                                             Re &                                           F1 \\
\midrule
     rotated &  \begin{tabular}{@{}c@{}}ALOOC \\ OOB \\ ODIN\end{tabular} &  
\begin{tabular}{@{}c@{}}0.0 \\ \textbf{0.09} \\ -0.12\end{tabular} &  
\begin{tabular}{@{}c@{}}\textbf{0.0} \\ 0.75 \\ 1.0\end{tabular} &  \begin{tabular}{@{}c@{}}0.55 \\ 0.16 \\ \textbf{0.02}\end{tabular} &  \begin{tabular}{@{}c@{}}\textbf{1.0} \\ 0.21 \\ 0.19\end{tabular} &  \begin{tabular}{@{}c@{}}0.45 \\ 0.84 \\ \textbf{0.98}\end{tabular} &  \begin{tabular}{@{}c@{}}\textbf{0.62} \\ 0.38 \\ 0.06\end{tabular} \\

\midrule

snow  (5) &  \begin{tabular}{@{}c@{}}ALOOC \\ OOB \\ ODIN\end{tabular} &  
\begin{tabular}{@{}c@{}}\textbf{0.81} \\ 0.01 \\ 0.16\end{tabular} &  
\begin{tabular}{@{}c@{}}\textbf{0.0} \\ 0.83 \\ 0.85\end{tabular} &  \begin{tabular}{@{}c@{}}0.29 \\ 0.16 \\ \textbf{0.02}\end{tabular} &  \begin{tabular}{@{}c@{}}\textbf{1.0} \\ 0.2 \\ 0.22\end{tabular} &  \begin{tabular}{@{}c@{}}0.71 \\ 0.84 \\ \textbf{0.98}\end{tabular} &  \begin{tabular}{@{}c@{}}\textbf{0.94} \\ 0.29 \\ 0.28\end{tabular} \\

\midrule

fog  (5) &  \begin{tabular}{@{}c@{}}ALOOC \\ OOB \\ ODIN\end{tabular} &  
\begin{tabular}{@{}c@{}}-0.28 \\ \textbf{0.0} \\ \textbf{0.0}\end{tabular} &  
\begin{tabular}{@{}c@{}}0.93 \\ \textbf{0.84} \\ 0.98\end{tabular} &  \begin{tabular}{@{}c@{}}0.29 \\ 0.16 \\ \textbf{0.02}\end{tabular} &  \begin{tabular}{@{}c@{}}0.16 \\ \textbf{0.2} \\ \textbf{0.2}\end{tabular} &  \begin{tabular}{@{}c@{}}0.71 \\ 0.84 \\ \textbf{0.98}\end{tabular} &  \begin{tabular}{@{}c@{}}0.15 \\ \textbf{0.28} \\ 0.1\end{tabular} \\
\bottomrule
\end{tabular}
}
\end{table}

%% file: sections/5.tex
\section{Conclusion and research directions}\label{section-5}

In this work, we proposed a framework for benchmarking safety monitors for ML classifiers.
We argue that there is a need for a better framework for benchmark SMs based on data by applying more metrics beyond accuracy and AUROC curves.
Thus, we proposed a minimal set of measurements that should be considered when benchmarking such solutions.

Besides, we also showed that measuring these SMs is less straightforward than it seems. We presented two dimensions that need to be investigated: the SM’s overall impact on the SUT; and the detection performance of the SMs.
Our approach allowed insights into three categories of data-based monitors.

Our results indicate at least four general takeaways for the current solutions that are solely based on DNN’s data:
\begin{itemize}
    \item The overall accuracy in the detection tends to be as bad as a random classifier due to the high amount of false positives or false negatives, depending on the scenario.
    \item Such methods tend to negatively affect the system under test since they yield too many false positives interfering with the correct ML decision.
    \item They need a considerable amount of memory or disk space since a monitor can consume up to 100MB.
    \item They are not so fast to perform the detection and have a considerable overhead compared to the ML software.
\end{itemize}

The tested SM were exposed just to novelty detection in their original papers.
However, testing these SM to other four types of OOD data was important to show that:
\begin{itemize}
    \item \emph{SM based on DNN inputs}: it has the advantage that it does not need to inspect the internals of the DNN.
    Thus, it can make the monitoring independent of the ML classifier. 
    However, it has the drawback of being biased to the data used during the training.
    \item \emph{SM based on DNN intermediate values}: it has the advantage of inspecting the ML model as a white box.
    That is, it allows to look at the internal values that led the DNN to decide. 
    However, the drawback is that the activation functions are insufficient to provide all information helpfully learned from the DNN.
    \item \emph{SM based on DNN outputs}: it has the advantage that it decreases and equilibrates the confidence values from the DNN. 
    However, similarly to approaches based on DNN intermediate values, the drawback is that it relies heavily on the performance of the DNN.

\end{itemize}
 
A limitation of this work is that we did not apply more and bigger DNN architectures. Such experiments could produce a more robust evaluation.
Moreover, we benchmarked just over image datasets, not considering an entire system.
Therefore, it is also essential to investigate the performance of the SMs inside a simulator to achieve a complete analysis.
Finally, as future work, we also want to add SMs based on uncertainty quantification \cite{berend2020cats}, \cite{weiss2021fail} to our experiments.

The main conclusion is that using such data-based monitors does not provide sufficient confidence for their use in a safety-critical application.
Thus, we want to explore how to combine data-based and model-based monitors.